\documentclass[sn-apa]{sn-jnl}

\usepackage{bm} 
\usepackage[T1]{fontenc}
\usepackage{tipa} 
\usepackage{textcomp} 
\usepackage{bookmark} 
\usepackage{booktabs}
\usepackage{array}
\usepackage{siunitx}
\usepackage[normalem]{ulem} 
\usepackage{wrapfig}
\usepackage{calc}
\usepackage{qrcode}
\usepackage{etoolbox}
\newsavebox\captionqr

\renewcommand*{\backref}[1]{}
\renewcommand*{\backrefalt}[4]{\small{\ifcase #1 Not cited.%
          \or Cited on page~#2.%
          \else Cited on pages #2.%
    \fi%
    }}

\jyear{2021}
\theoremstyle{thmstyleone}%
\theoremstyle{thmstyletwo}%
\theoremstyle{thmstylethree}%
\begin{document}
\title{Making sense of spoken plurals}
\author*[1]{\fnm{Elnaz} \sur{Shafaei-Bajestan}}\email{elnaz.shafaei-bajestan@uni-tuebingen.de}
\author[2]{\fnm{Peter } \sur{Uhrig}}\email{peter.uhrig@fau.de}
\author[1]{\fnm{R. Harald} \sur{Baayen}}\email{harald.baayen@uni-tuebingen.de}
\affil[1]{\orgname{Eberhard Karls Universität Tübingen}, \country{Germany}}
\affil[2]{\orgname{Friedrich-Alexander-Universität Erlangen-Nürnberg}, \country{Germany}}

\newpage
\abstract{
Distributional semantics offers new ways to study the semantics of morphology.  This study focuses on the semantics of noun singulars and their plural inflectional variants in English. Our goal is to compare two models for the conceptualization of plurality. One model (FRACSS) proposes that all singular-plural pairs should be taken into account when predicting plural semantics from singular semantics. The other model (CCA) argues that conceptualization for plurality depends primarily on the semantic class of the base word. 
We compare the two models on the basis of how well the speech signal of plural tokens in a large corpus of spoken American English aligns with the semantic vectors predicted by the two models. Two measures are employed: the performance of a form-to-meaning mapping and the correlations between form distances and meaning distances. 
Results converge on a superior alignment for CCA. 
Our results suggest that usage-based approaches to pluralization in which a given word's own semantic neighborhood is given priority outperform theories according to which pluralization is conceptualized as a process building on high-level abstraction. 
We see that what has often been conceived of as a highly abstract concept, [+\textsc{plural}], is better captured via a family of mid-level partial generalizations.}



\maketitle

\medskip

\newpage

\section{Introduction}\label{sec:introduction}

\begin{figure}[htbp]
\includegraphics[width=0.9\textwidth]{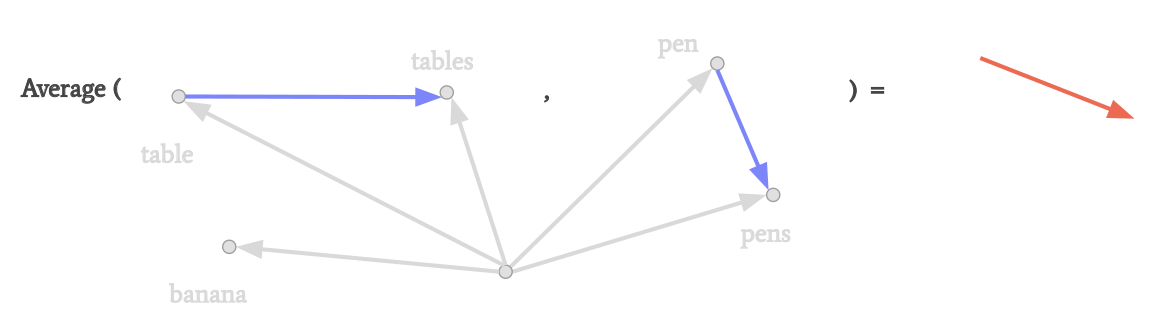} \\
\hspace{2em}\includegraphics[width=0.7\textwidth]{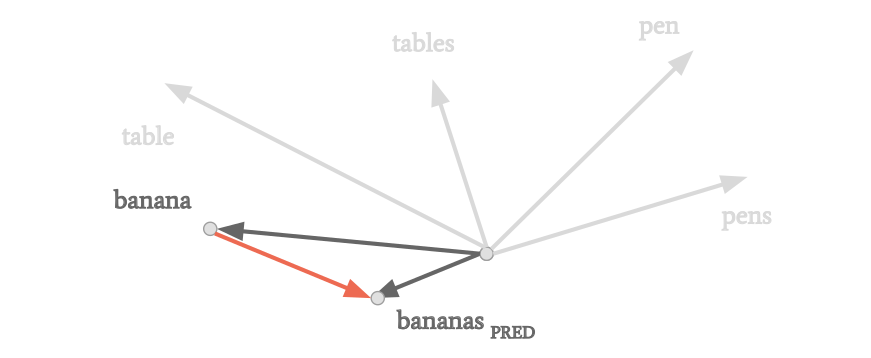} \\

\caption{The average of the shift vectors for given singular-plural pairs (\textit{table/tables}, \textit{pen/pens}) is used to calculate the semantic vector of the unknown plural vector of \textit{banana}.}
\label{fig:bananas}
\end{figure}

{\color{black}
Distributional semantics \citep{Landauer:Dumais:1997,Firth:1968,Harris1954DistributionalStructure,Mikolov:Chen:Corrado:Dean:2013,Wang:Wang:Chen:Wang:Kuo:2019,Gunther:Rinaldi:Marelli:2019,Boleda:2020} offers new opportunities for understanding the semantics of word formation \citep[see, e.g.][]{Marelli:Baroni:2015,Kisselew:Pado:Palmer:Snajder:2015} and inflection \citep[see, e.g.,][]{Baayen:Moscoso:2005}.  Given corpus-based semantic vectors (known as embeddings in computational linguistics and natural language processing) for pairs of base words and corresponding complex words, several methods have been proposed that take as input the semantic vector of the base word, and that produce as output the vector of the complex word.  One such method is illustrated in Figure~\ref{fig:bananas}. Given the semantic vectors for two pairs of singulars and plurals (\textit{table/tables} and \textit{pen/pens}), and given the semantic vector for \textit{banana} but no semantic vector for its plural, the semantic vector for \textit{bananas} is obtained by first calculating the vectors that start at a singular and point to the corresponding plural (represented by blue vectors), and average these, resulting in an average shift vector (in red).  This shift vector can then be applied to the vector of \textit{banana}, resulting in the semantic vector for \textit{bananas} (lower panel). \citet{Kisselew:Pado:Palmer:Snajder:2015} calculated the average shift vector for each of a large set of German derivational affixes, and showed that this results in high-quality estimates of the meanings of derived complex words. \citet{Marelli:Baroni:2015} used a method based on matrix multiplication to obtain predicted semantic vectors for derived words, and showed that this method generated quantitative predictors that help explain variance in measures of lexical processing such as reaction times in visual lexical decision.} 

{\color{black}
The abovementioned studies associate each derivational exponent with one conceptualization function that predicts the complex word from its base word. Recent research on plural inflection has clarified that the semantics of plurality can be too complex semantically to be represented by a single shift vector. \citet{chuang:2022} observed that for Russian, a fusional language, the shift vector for a singular to a plural noun embedding varies systematically by each of the six cases. \citet{nikolaev:2022} report that also for Finnish, an agglutinative language, plural shift vectors are different for each of the 14 cases.  Furthermore, \citet{Shafaei:Tari:Uhrig:Baayen:2022:Morphology} report that in English, shift vectors are conditioned on the semantic class of the base word.
}

\begin{figure}[b]
    \centering
    \includegraphics[width=\textwidth]{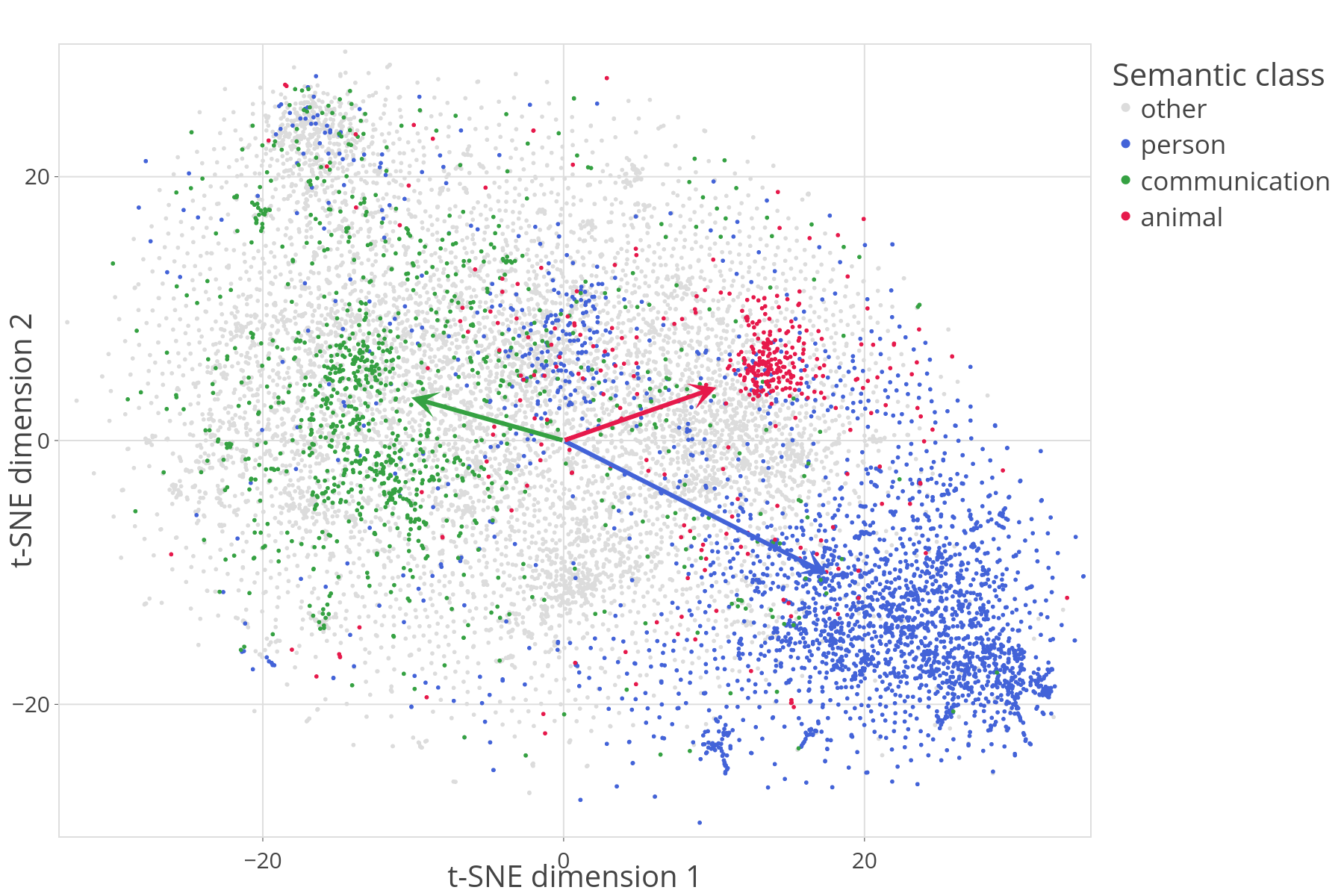}
    \caption{A projection of shift vectors \citep[calculated from {\tt word2vec} embeddings,][]{Mikolov:Chen:Corrado:Dean:2013} onto a two-dimensional plane using the t-SNE dimension reduction algorithm reveals semantic clustering ($N=11749$). {\color{black} Each dot represents a shift vector, one for each singular-plural pair. Three semantic classes (based on WordNet) are highlighted: animals (red), persons (blue), and communication words (green). If the semantics of pluralization would be identical for all words, a single cluster with its centroid at a distance from the origin would have emerged from the analysis.}
    An interactive version of this plot can be found \href{https://quantling.org/MentalLexicon/English/English-shift-space-w2v-3categories.html}{here}.}
    \label{fig:morphology_teaser}
\end{figure}

Figure~\ref{fig:morphology_teaser} illustrates this phenomenon, highlighting three clusters of shift vectors in a two-dimensional plane constructed with the t-SNE dimension reduction technique \citep{Maaten:Hinton:2008}.  T-SNE is designed such that if observations cluster in a high-dimensional space, these clusters are (with high probability) optimally visible and distinct in its two-dimensional re-representation of this space.  In Figure~\ref{fig:morphology_teaser},  words denoting people (e.g., \textit{president}) are found predominantly in the lower right (in blue), words for communication (e.g., \textit{language}) cluster in the center-left (in green), and words for animals (e.g., \textit{rabbit}) are found in the upper right of the t-SNE plane (in red).  The grey dots  represent nouns that belong to other semantic classes. {\color{black}  \citet{Shafaei:Tari:Uhrig:Baayen:2022:Morphology} propose to break down the set of nouns into some 400 semantic classes, derived from WordNet \citep{Miller:1995}. In this way, overlap between clusters is reduced, and plural vectors constructed by adding a class-specific shift vector to the vector of the singular are more similar to the empirical vectors. Class-specific average vectors are visualized with arrows in Figure 1 for the three example clusters.}

{\color{black} We note here that in a t-SNE plane constructed for singular and plural embeddings, plurals tend to occur very close to their singulars, and semantically similar words tend to be relatively close together.  What the t-SNE analysis of the shift vector clarifies is that how plural vectors are positioned with respect to their singular vectors varies considerably and systematically by semantic class.}

In English, words belonging to a semantic class are not overtly marked by a specific exponent.  Interestingly, there are languages that have grammaticalized class-conditional plural semantics.  Swahili, a Bantu language, makes use of a large number of noun classes, many of which are semantically motivated \citep{polome1967swahili}.  For instance, the m-wa class is used for persons, the ma-ji class comprises words for objects that occur in clusters such as fruits, and the ki-vi class covers artifacts and tools.  {\color{black} The Kiowa language from the Tanoan language family, has nine semantically motivated noun classes  \citet{Harbour:2008:MorphosemanticNumber,Harbour:2011}.  In contrast to English, Swahili and Kiowa have grammaticalized interactions of semantic class and plurality, indicating that speakers of these languages have made explicit in their grammars some perceived differences in what it means to be a plural.  
}

Further evidence for the grammaticalization of number by semantic class is provided by languages with classifiers such as Mandarin Chinese \citep[see, e.g.][]{yip2006chinese}. Constructions with numerals such as `two snakes' require a classifier before the noun, as in \textit{\`{e}r t\'{\i}ao sh\'{e}}, `two \textsc{classifier} snake'. The classifier \textit{tiao} is often found preceding long and flexible objects, such as snakes and rivers. Mandarin classifiers thus group multiples of objects together by their shape, size, or classes of artifacts (such as vehicles), among many others.
 
\begin{figure}[t]
\centering
\begin{minipage}[h]{0.45\textwidth}
\includegraphics[width=1.0\textwidth]{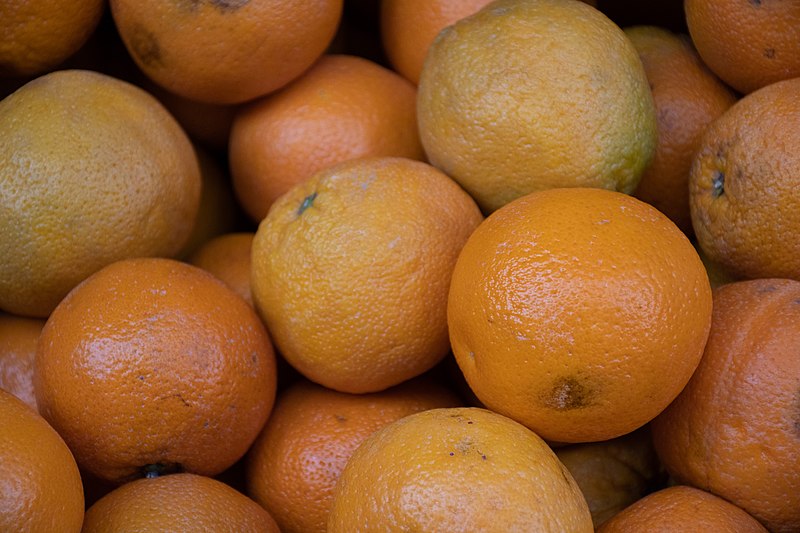} 
\end{minipage} \hspace{0.02\textwidth}
\begin{minipage}[h]{0.45\textwidth}
\includegraphics[width=1.0\textwidth]{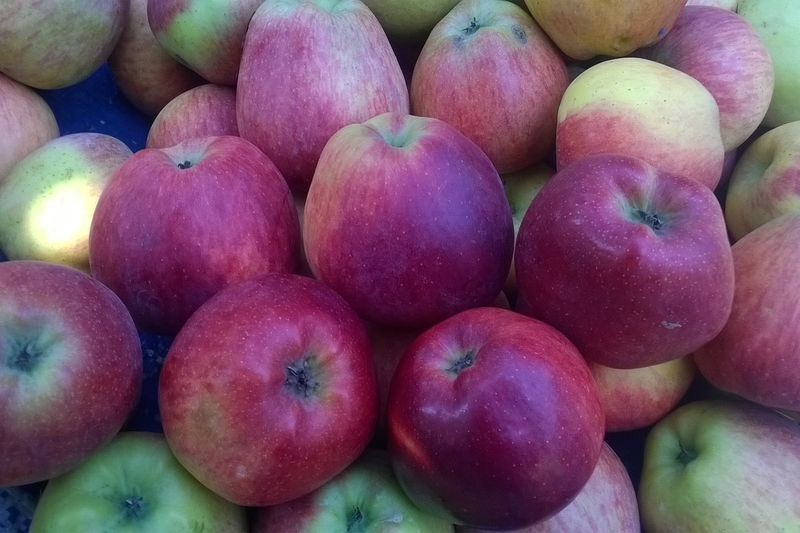} 
\end{minipage}

\vspace{0.03\textwidth}

\begin{minipage}[h]{0.45\textwidth}
\includegraphics[width=1.0\textwidth]{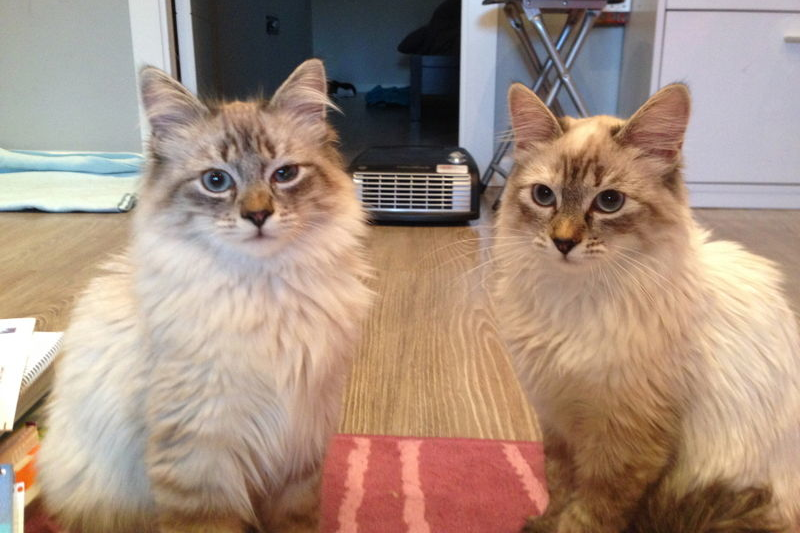} 
\end{minipage} \hspace{0.02\textwidth}
\begin{minipage}[h]{0.45\textwidth}
\includegraphics[width=1.0\textwidth]{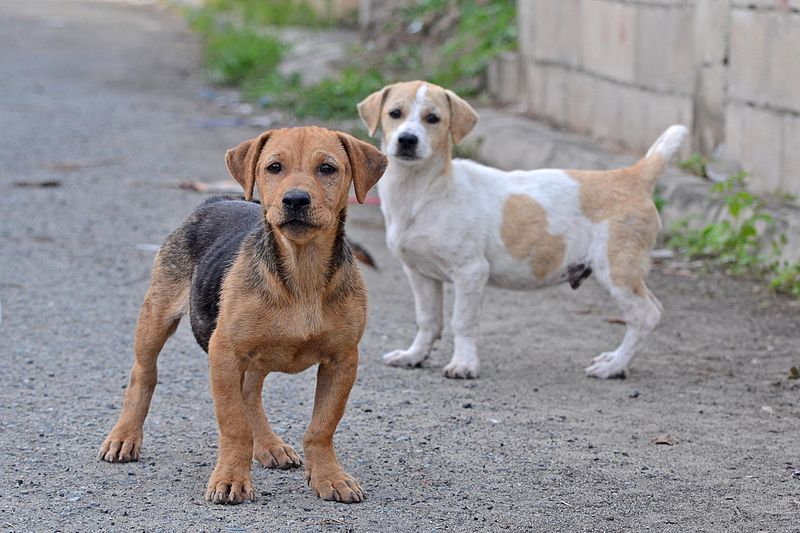} 
\end{minipage}

\caption{Multiples of different kinds of objects typically occur in different configurations. The images for oranges and dogs are obtained without modification from \cite{oranges} and \cite{dogs}. The images for apples and cats, obtained from \cite{apples} and \cite{cats}, have been cropped.}
\label{fig:apples}
\end{figure}

{\color{black}
Grammaticalization of number by semantic class is perhaps unsurprising as for many concrete plural nouns,  referents tend to occur in markedly different configurations. This is illustrated in Figure~\ref{fig:apples}. Oranges and apples appear in groupings that are entirely different from the configurations in which cats and dogs are typically found. It is remarkable that configurational properties of multiples of objects are so well captured by  word embeddings.
}

{\color{black}
The Kiowa noun system led \citet{Harbour:2008:MorphosemanticNumber,Harbour:2011} to argue for a morphosemantic theory of number.  The studies by \citet{nikolaev:2022}, \citet{chuang:2022}, and \citet{Shafaei:Tari:Uhrig:Baayen:2022:Morphology} provide first steps in the formulation of such a theory.  Focusing on noun inflection in English, the finding that plural semantics vary systematically by semantic class raises the question of how to formalize the conceptualization process that underlies the use of singular and plural nouns in English.  Given a semantic vector $\bm{v}_{\text{sg}}$ for the singular of a given lexeme, we can ask what kind of conceptualization operation $\Phi$ is needed to map this singular onto the corresponding plural $\bm{v}_{\text{pl}}$:
\begin{equation}
\Phi(\bm{v}_{\text{sg}}) = \bm{v}_{\text{pl}}.
\end{equation}
\citet{Shafaei:Tari:Uhrig:Baayen:2022:Morphology} investigated two alternative implementations of $\Phi$. The first implementation straightforwardly builds on the observed clustering in the shift vector space. 
\begin{equation}
\Phi(\bm{v}_{\text{sg}})  =  \bm{v}_{\text{sg}} + \bm{v}_{\text{shift} \mid \text{semantic class}}.
\end{equation}
For plurals that are `known' and for which semantic vectors are available, a plural vector is semantically decomposed in the following way:
\begin{equation} \label{eq:cca_e}
  \bm{v}_{\textsc{pl}} = \bm{v}_{\text{sg}} + \bm{v}_{\text{shift} \mid \text{semantic class}} + \bm{\epsilon}.
\end{equation}
In other words, known plurals are composed of the singular semantic vector, a shift vector appropriate to the semantic class of their base words, and a residual error vector that specifies the semantics that are specific to a given plural form.   As first pointed out by \citet{Sinclair:1991}, regular inflected words often have their own collocational profiles, which are part of native speakers' knowledge of the lexis of their language \citep[see also][]{Milin:etal:2009,Moscoso:Kostic:Baayen:2004}.  In (\theequation), this word specific knowledge that is not predictable from the semantic class is represented by the error vector $\bm{\epsilon}$. 

For `out-of-vocabulary' plurals, there is no error vector, and the semantic vector predicted for these plurals simplifies to:
\begin{equation}
  \bm{v}_{\textsc{pl}} = \bm{v}_{\text{sg}} + \bm{v}_{\text{shift} \mid \text{semantic class}}.\label{eq:cca0}
\end{equation}
This equation can also be used when evaluating how well the conceptualization function $\Phi$ generalizes to data it has not been trained on.  This approach, which applies the approach to derivation of \citet{Kisselew:Pado:Palmer:Snajder:2015} and extends it to the English plural, was laid out in \citet{Shafaei:Tari:Uhrig:Baayen:2022:Morphology}. Following their terminology, we henceforth refer to this method as Cosine Class Average (CCA). 
}


An alternative approach to plural conceptualization builds on the FRACSS model of \citet{Marelli:Baroni:2015}, which was originally developed for derivational morphology.  When all singular semantic vectors are brought together as the row vectors of a matrix $\bm{S}$, and all corresponding plural vectors are bundled together as the row vectors of a matrix $\bm{P}$, a linear mapping $\bm{M}$ can be obtained by solving
$$
\bm{S}\bm{M} = \bm{P}.
$$
Given $\bm{M}$, the conceptualization function $\Phi$ is given by
\begin{equation}
\Phi(\bm{v}_{\textsc{sg}}) = \bm{v}_{\textsc{sg}}\bm{M},
\end{equation}
the embedding of a plural is decomposed as
\begin{equation} \label{eq:fracss_e}
\bm{v}_{\textsc{pl}} = \bm{v}_{\textsc{sg}}\bm{M} + \bm{\epsilon}, 
\end{equation}
and for plurals for which an embedding is not available, the plural vector is estimated by
\begin{equation}
\bm{v}_{\textsc{pl}} = \bm{v}_{\textsc{sg}}\bm{M}. \label{eq:fracss0}
\end{equation}


The conceptualization functions $\Phi_{\textsc{cca}}$ and 
$\Phi_{\textsc{fracss}}$ implement two different ways of thinking about the conceptualization of plural nouns. The perspective taken by $\Phi_{\textsc{fracss}}$ is global in nature: to predict a plural meaning, one needs to take into account the systematicities that govern all singular and plural pairs.  By contrast, the perspective taken by $\Phi_{\textsc{cca}}$ is local in nature, the hypothesis being that it is only the systematicities between singulars and plurals within the semantic class to which the lexeme belongs that are relevant. 

However, although $\Phi_{\textsc{cca}}$ straightforwardly implements the structure  that is revealed by the t-SNE analysis to be present in the high-dimensional space of embeddings, precisely because it builds on semantic classes, it is actually a complex theory: semantic classes are not straightforward to define.  FRACSS, by contrast, is a theory that takes as its point of departure that there is a single plural operation. Because linear transformations are quite powerful, it is possible that the linear transformation of FRACSS will actually capture in its stride the semantic conditioning that is revealed by the t-SNE analysis.

\citet{Shafaei:Tari:Uhrig:Baayen:2022:Morphology} observed that the plural vectors predicted by $\Phi_{\textsc{fracss}}$ and $\Phi_{\textsc{cca}}$ are in general very similar. The former plural vectors were somewhat more similar to the word2vec gold standard plural vectors in terms of angle, the latter vectors were somewhat more similar in terms of vector length. 
{\color{black} 
In order to assess the relative merits of the two conceptualization functions, \citet{Shafaei:Tari:Uhrig:Baayen:2022:Morphology} therefore investigated the relation between plurals' semantics and their forms.  Different approaches to clarifying this relation have been proposed in the literature. \citet{Tamariz:2008, Shillcock:Kirby:McDonald:Brew:2001, Monaghan:Shillcock:Christiansen:Kirby:2014} compared distances in the form space to distances in a distributional semantic space.  \cite{Levy:Kenette:Oxenberg:Castro:DeDeyne:Vitevitch:Havlin:2021} employed methods from network science to build a multi-layer network that connects a phonological network to a semantic network.  Lexical measures such as orthography-semantics consistency \citep{Marelli:Amenta:Crepaldi:2015, Siegelman:Rueckl:Lo:Kearns:Morris:Compton:2022} and phonology-semantics consistency \citep{Amenta:Marelli:Sulpizio:2017} are now available for capturing the effects of form-meaning associations on lexical processing.

\citet{Shafaei:Tari:Uhrig:Baayen:2022:Morphology} addressed the systematicities between form and meaning of English plural nouns using the Discriminative Lexicon Model (DLM) proposed by  \citet{Baayen:Chuang:Shafaei:Blevins:2019}.  This computational model approximates comprehension with a linear mapping from numeric form representations to semantic vectors.  A core assumption of the DLM is that the form space and the semantic space are so similar and well calibrated to each other that simple linear mappings are sufficient for obtaining high-quality (albeit not perfect) predictions for meanings given forms (in comprehension), and for forms given meanings (in production).  Given that according to this model the mappings between form and meaning are extremely simple, the quality of the model's predictions is critically dependent on the representations of  form and of meaning.  Within this modeling framework, it is therefore essential to clarify whether the semantic vectors of plurals are better aligned with words forms when generated with FRACSS, or when generated with CCA.
}
Using form representations that encode which triphones are present in a word's form, \citet{Shafaei:Tari:Uhrig:Baayen:2022:Morphology} found that plural vectors obtained with $\Phi_{\textsc{cca}}$ could be predicted more precisely from their forms than plural vectors obtained with $\Phi_{\textsc{fracss}}$, especially when the model was tested on held-out data.

A drawback of using form vectors that are derived from (tri)phones is that phone-based representations do not do justice to both the richness and the enormous variability of  spoken words.  The goal of the present study is to clarify how well $\Phi_{\textsc{cca}}$ and $\Phi_{\textsc{fracss}}$ are aligned with the audio signal. 

In what follows, we report two experiments addressing this issue.  The first experiment follows \citet{Shafaei:Tari:Uhrig:Baayen:2021} and makes use of a version of the Discriminative Lexicon Model that was developed specifically for auditory comprehension: the LDL-AURIS model \citep{Shafaei:Tari:Uhrig:Baayen:2021}.  As this model derives form vectors for word tokens from the audio files of these tokens, it is well-suited to address the relation between specifically words' speech audio and their meanings.

%
%
%

The second experiment follows in the footsteps of \citet{Tamariz:2008, Shillcock:Kirby:McDonald:Brew:2001, Monaghan:Shillcock:Christiansen:Kirby:2014}. These studies make use of distance measures, which for words' forms are based on orthographic or phone-based representations. In section~\ref{sec:repr}, we study the distances of form vectors that we extracted from the audio files of word tokens, and compare these with the distances of semantic vectors.  Our question of interest is whether distances in `audio space' are more similar to distances in semantic space when plural meanings are estimated with $\Phi_{\textsc{fracss}}$ or with $\Phi_{\textsc{cca}}$.


In what follows, we first introduce the dataset on which our analysis is based. We then report the study using LDL-AURIS (section~\ref{sec:auris}), followed by the study using distances (section~\ref{sec:euclid}). We conclude with a discussion in section~\ref{sec:discussion}.  

\section{Data}\label{sec:data}

Our data are taken from the NewsScape English Corpus 2016 \citep{Uhrig:2018, Uhrig:Habil}.
This corpus is based on the US-American English-language recordings in the UCLA Library Broadcast NewsScape. The corpus contains TV news, including traditional newscasts and political talk shows as well as late-night shows and daytime talk shows. The vast majority of the language recorded can thus be assumed to be Standard English with a Network English accent. Overall, male speakers contribute much more than half of the words. In a sample of shows from broadcaster CNN for which transcripts were available online, even 80\% of the words were uttered by male speakers \citep{Uhrig:Habil}. In terms of age distribution, younger speakers, especially children and teenagers, are underrepresented. The initial audio quality is usually quite good because much of the audio comes from recording studios, but the corpus also includes footage from outside reporters. There are also programs with audiences, who cheer and applaud, all reducing the quality of the audio signal.

For the year 2016, these recordings amount to more than 35,000 hours with around 269 million tokens of subtitles. Directly after subtitle extraction, the recordings are compressed to 240 MB/hour, using H.264 for the video stream and the Fraunhofer FDK library to produce an AAC audio stream at 96 kbit/sec. The subtitles were processed with Stanford CoreNLP \citep{Manning:etal:2014:CoreNLP} version 3.7.0, i.e. annotated for part of speech with the Penn Treebank tagset using CoreNLP's caseless model. 

Subtitles are not intended to be accurate transcripts of the words spoken. For the sake of readability, they usually omit false starts, often leave out words and phrases --- in particular in fast dialogue --- and are not designed to cope with overlaps. Commercials often do not have subtitles at all. We thus expect a certain error in the forced alignment, which is designed to align a transcript with the corresponding audio recording. Due to the transcript quality and the length of the recordings, most forced alignment systems are not suitable for this use case. A slightly modified version of Gentle \citep{Ochshorn:Hawkins:2015:Gentle}, which markets itself as a ``robust yet lenient forced aligner built on Kaldi''\footnote{https://lowerquality.com/gentle/}, was deployed on the corpus. 
While the software reported a success rate of somewhere between 90\% and 95\% on average, we were able to select for inclusion only those files where Gentle's self-reported success rate was at least 97\% because we only needed a much smaller dataset than the entire corpus.

From this well-aligned part of the corpus, we selected 500 hours of programs and extracted a total of 750,816 singular and plural audio word tokens of all 2062 orthographic word form types. The types were selected with the constraints that their frequency was at least 70 in the 500-hour corpus and that a word2vec semantic vector was available.
Proper names, plurals endings with anything other than an \textit{-s}, plural-singular pairs with the same word form, and named entities were excluded from the dataset. 

The phonemic transcriptions that are provided in the NewsScape English Corpus 2016 were obtained with the Gentle forced aligner \citep{Ochshorn:Hawkins:2015:Gentle}. Gentle runs an automatic speech recognition process with a bigram language model created from the transcript in the background. It is based on Kaldi ASR \citep{Povey:etal:2011:Kaldi}, which uses a purpose-built version of the CMUDict machine-readable pronunciation dictionary (\url{https://github.com/cmusphinx/cmudict}) that comes without stress information. Note that our dataset inherits the pronunciation variants offered by CMUDict, i.e. the word \textit{either} will be represented either as [a\textsci \dh \textschwa r] or as [i:\dh \textschwa r], depending on the pronunciation of the speaker. However, not all pronunciation variants typically found in American English are listed in the version of CMUDict used here. Thus \textit{don't} is transcribed as [do\textupsilon nt] or as [do\textupsilon n], but vowel reductions that are typical of fast speech in combinations such as \textit{don't know}, often rendered orthographically as \textit{dunno}, are not covered. Thus we have to be aware that while our dataset offers some variation in pronunciation, which makes it more ecologically valid than purely dictionary-based approaches without speech recognition, it will contain instances where the transcript does not correspond to the spoken form because the transcription of the spoken form was not available in CMUDict. 

\section{Study 1: assessing the isomorphy of form and meaning with LDL-AURIS}\label{sec:auris}

\subsection{Representations\label{sec:repr}}


For our investigations into the relation between spoken noun singulars and plurals and their semantics, we set up three sets of 300-dimensional semantic vectors. 
The first set of vectors based on pre-trained word2vec \citep{Mikolov:Chen:Corrado:Dean:2013} embeddings (for both singulars and plurals) was retrieved from \cite{googleWord2Vec}.\footnote{  In this study,  we accept the embeddings of word2vec as ground truth. A question we leave to further research is whether multi-modal embeddings \citep[see, e.g.,][for olfactory, auditory, and visual grounding of embeddings]{shahmohammadi2021learning,Kiela:Clark:2017,Kiela:Bulat:Clark:2015} will help enhance the modeling of the conceptualization of plurality in English.}
For the compilation of the other two sets, the vectors for singular words are identical to those of the first set.  For set 2, the plural vectors were calculated  according to $\Phi_{\textsc{cca}}$, and for set 3, according to $\Phi_{\textsc{fracss}}$.   The semantic classes used by the $\Phi_{\textsc{cca}}$ conceptualization function are taken from WordNet \citep{Ciaramita:Johnson:2003, Fellbaum:1998,Miller:1995}. For further details on the granularity of the conceptual classes used, see \citet{Shafaei:Tari:Uhrig:Baayen:2022:Morphology}.

\begin{figure}[bht]
    \savebox\captionqr{\qrcode[hyperlink,height=1.5cm]{http://go.redhenlab.org/pgu/0120}}
    \centering
    \includegraphics[height=9.2cm]{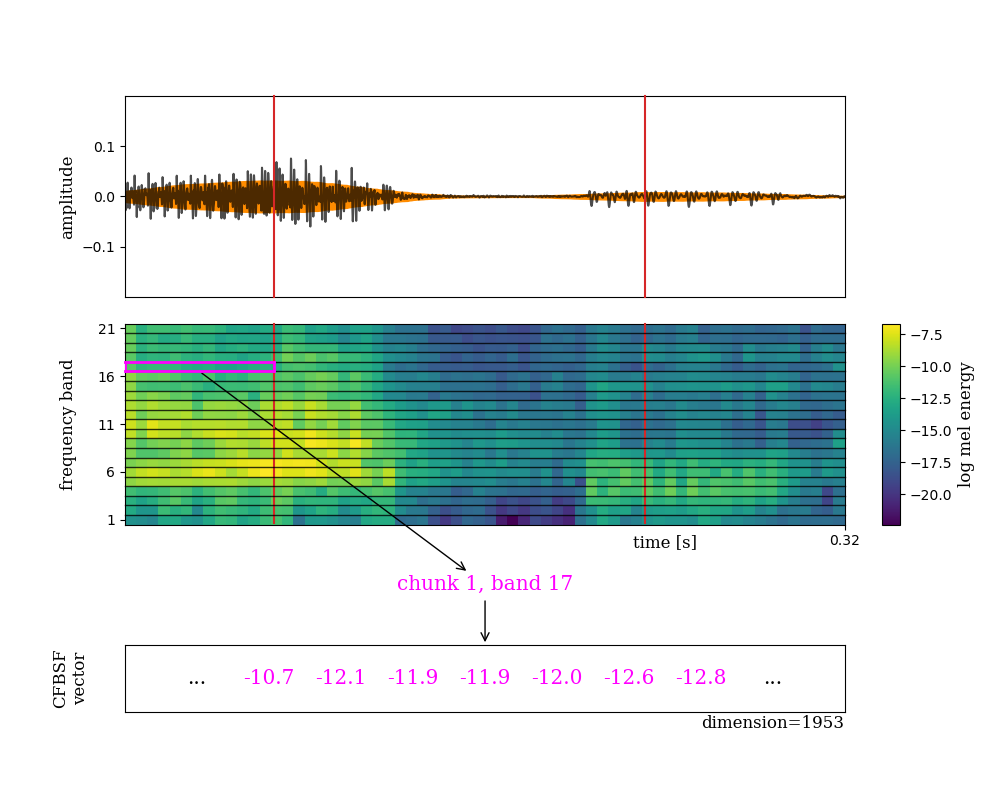}
    \llap{\raisebox{7.2cm}{\usebox\captionqr}}
    \caption{C-FBSF extraction for the audio of  rendition of the word \textit{apple} (duration of \SI{0.32}{\second}). \href{http://go.redhenlab.org/pgu/0120}{Click here} or scan the QR code to listen to the audio.  The top panel presents the waveform in dark gray, and the smoothed Hilbert amplitude envelope of the signal in orange. The red vertical lines indicate chunk boundaries, located at the local maxima of the Hilbert envelope at \SI{0.07}{\second} and \SI{0.23}{\second}. The mid panel presents the corresponding log MEL spectrogram (using an FFT window size of \SI{0.005}{\second}) that depicts log energy values as colors over time (shown on the x-axis) split at 21 auditory filter banks according to the MEL scale (shown on the y-axis). The upper and mid panels are time-aligned. Lower panel: a segment of the output C-FBSF vector that corresponds to the first 7 sampled energy values from the 17th frequency band of the first chunk. The length of the C-FBSF vector compiled from all three chunks and 21 frequency bands before zero-padding is 1953.}
    \label{fig:cfbsf}
\end{figure}

For representing the speech signal of singular and plural nouns, we made use of the Continuous Frequency Band Summary Features (C-FBSF) developed by \cite{Shafaei:Tari:Uhrig:Baayen:2021}.  Figure~\ref{fig:cfbsf} illustrates, for one example audio token which can be accessed through the QR link in the upper right of the figure, how these features are calculated. The algorithm takes the digital signal of a word as input (top panel) and divides it into chunks based on the maxima of the periodic parts of the signal.   The MEL spectrogram (middle panel) for each chunk is divided into 21 frequency bands, mirroring the frequency bands of the cochlea. For each frequency band, an order-preserving random sample of length 20 is taken, and correlation coefficients of the values at the current frequency band with those of the following bands are calculated and included.  These numeric vectors are concatenated first within chunks and then between chunks (lower panel). 

\noindent
The dimensionality of the auditory space is determined by the number of chunks in the  longest word in the dataset, which is 4 for the present data. As a consequence, the auditory space is 8,463-dimensional. All C-FBSF vectors are zero-padded to match the 8,463 dimensions. For the audio token \textit{apple} portrayed in Figure~\ref{fig:cfbsf}, which has a duration of 0.32 seconds, an initial 1953-dimensional vector is thus padded with 6510 trailing zeros. 

\subsection{Mapping audio vectors onto semantic vectors}

The form vectors derived from the audio files were brought together as the row vectors of a form matrix $\bm{C}$. The corresponding semantic vectors were bundled as the row vectors of three semantic matrices $\bm{S}$: one with the  word2vec gold standard vectors, one replacing word2vec plural vectors with FRACSS vectors, and one replacing word2vec plural vectors with CCA plural vectors.   

LDL-AURIS calculates a mapping $\bm{F}$ from the form matrix $\bm{C}$ to a semantic matrix $\bm{S}$ by solving 
$$
\bm{CF}=\bm{S}. 
$$
We used the normal equations \citep{Faraway:2005} for the multivariate multiple regression model to estimate $\bm{F}$, using the Moore-Penrose algorithm \citep{Moore:1920} for matrix inversion.  Once $\bm{F}$ has been estimated, we obtain the predicted semantic matrix $\hat{\bm{S}}$ as follows:
\begin{equation}
\hat{\bm{S}} = \bm{C}\bm{F}.
\label{eq:SisCF}
\end{equation}
We write $\hat{\bm{S}}$ rather than $\bm{S}$ because, as in regression, the predicted row vectors $\hat{\bm{s}}$ of $\hat{\bm{S}}$ are not identical to the gold standard row vectors $\bm{s}$ of $\bm{S}$.  A specific predicted semantic vector $\hat{\bm{s}}_i$ is judged to be accurate when it is closer to the corresponding gold-standard vector $\bm{s}_i$ than to any other semantic vector $\bm{s}_j, j \neq i$.  As a measure of semantic proximity, LDL-AURIS uses the Pearson correlation. 

\subsection{Training data and test data}\label{subsec:data}

For evaluating the quality of a mapping $\bm{F}$, we split the data into training and test sets using a stratified 10-fold cross-validation design with respect to the word types. Cross-validation is a standard procedure in machine learning for model evaluation.  In 10-fold cross-validation, the dataset is split into 10 parts (henceforth folds).  Ten models are fitted on nine of the folds, and are tested on the remaining fold.
Stratification ensures that each fold is appropriately representative of the whole dataset. We made sure that all word types appear in every training and test set and that the relative frequencies of audio tokens for each word type are preserved across training and testing.  Table \ref{tab:data} reports the number of types and the number of tokens averaged over the 10 folds for singular and plural words, broken down by whether the corresponding plural or the corresponding singular appears in the training set.

\begin{table}[ht]
    \centering
    \caption{The number of word form types and the average number of audio tokens (and standard deviation) across 10 stratified folds.}
    \label{tab:data}
    \begin{tabular}{lrr}
    \toprule
                                 Dataset &  Word form types & Average audio tokens (SD) \\
    \midrule
                                   Train &            2062 &        675734 (1.26) \\
        \hspace{5mm}singular with plural &             475 &        273893 (6.26) \\
     \hspace{5mm}singular without plural &             979 &        243630 (9.10) \\
        \hspace{5mm}plural with singular &             499 &        141214 (7.95) \\
     \hspace{5mm}plural without singular &             109 &         16996 (4.81) \\
                                    Test &            2062 &         75082 (1.26) \\
        \hspace{5mm}singular with plural &             475 &         30433 (6.26) \\
    \hspace{5mm}singular without plural  &             979 &         27070 (9.10) \\
        \hspace{5mm}plural with singular &             499 &         15690 (7.95) \\
     \hspace{5mm}plural without singular &             109 &          1888 (4.81) \\
    \bottomrule
    \end{tabular}
\end{table}

\subsection{Procedure}

For evaluating the three k
! LaTeX Error: File `algorithm.sty' not found.inds of predicted plural semantic vectors, we constructed the gold standard semantic matrices $\bm{S}_{\text{\tiny{CCA}}}$, $\bm{S}_{\text{\tiny{FRACSS}}}$, and $\bm{S}_{\text{\tiny{word2vec}}}$, each of which contains row vectors for both singulars and plurals. The CCA and FRACSS vectors for plurals were estimated with equations (\ref{eq:cca0}) and (\ref{eq:fracss0}) respectively. Using the form matrix $\bm{C}$, we calculated three mappings  $\bm{F}_{\text{\tiny{CCA}}}$,  $\bm{F}_{\text{\tiny{FRACSS}}}$, and $\bm{F}_{\text{\tiny{word2vec}}}$. Applying  (\ref{eq:SisCF}), we then obtained the estimated vectors $\hat{\bm{S}}_{\text{\tiny{CCA}}}$, $\hat{\bm{S}}_{\text{\tiny{FRACSS}}}$, and $\hat{\bm{S}}_{\text{\tiny{word2vec}}}$.

It is worth keeping in mind that the rows of the form matrix $\bm{C}$ are all distinct, and that the total number of rows in the $\bm{C}$ matrix equals the number of audio tokens in our dataset.  By contrast, many rows of the semantic matrix $\bm{S}$ are duplicated, with as many duplicates for a given noun type as there are tokens for that noun type. As the estimated semantic vectors are \textit{estimates}, the row vectors of $\hat{\bm{S}}$ are again all unique. However, if the mappings are of good quality, the estimated semantic vectors for the audio vectors of the same word type are expected to be highly similar. 

We also note that because mappings are estimated for singulars and plurals jointly, changing the semantic vectors of plurals will affect the semantic vectors predicted for singulars. In other words, even though the form and meaning vectors of singulars are identical across the three datasets, the estimated singular vectors will be somewhat different, depending on the quantitative properties of the plural semantic vectors.

We evaluated the quality of the estimated semantic vectors in two ways. First, we calculated accuracy on both the training data and the test data.  The top-$N$ accuracy of the model is defined as the percentage of the test items for which we find the predicted vector among the top-$N$ vectors with the strongest correlations with the targeted output vector.

Since the word types have highly unequal frequencies with a Zipf-like probability distribution, we also used the $F_1$ score (the harmonic mean of precision and recall) to gauge the quality of the mappings $\bm{F}$ and the different semantic vectors for plurals that these mappings predict.

\subsection{Results}

\subsubsection{Recognitions}
! LaTeX Error: File `algorithm.sty' not found.

\begin{figure}[ht]
    \centering
    \includegraphics[width=\textwidth]{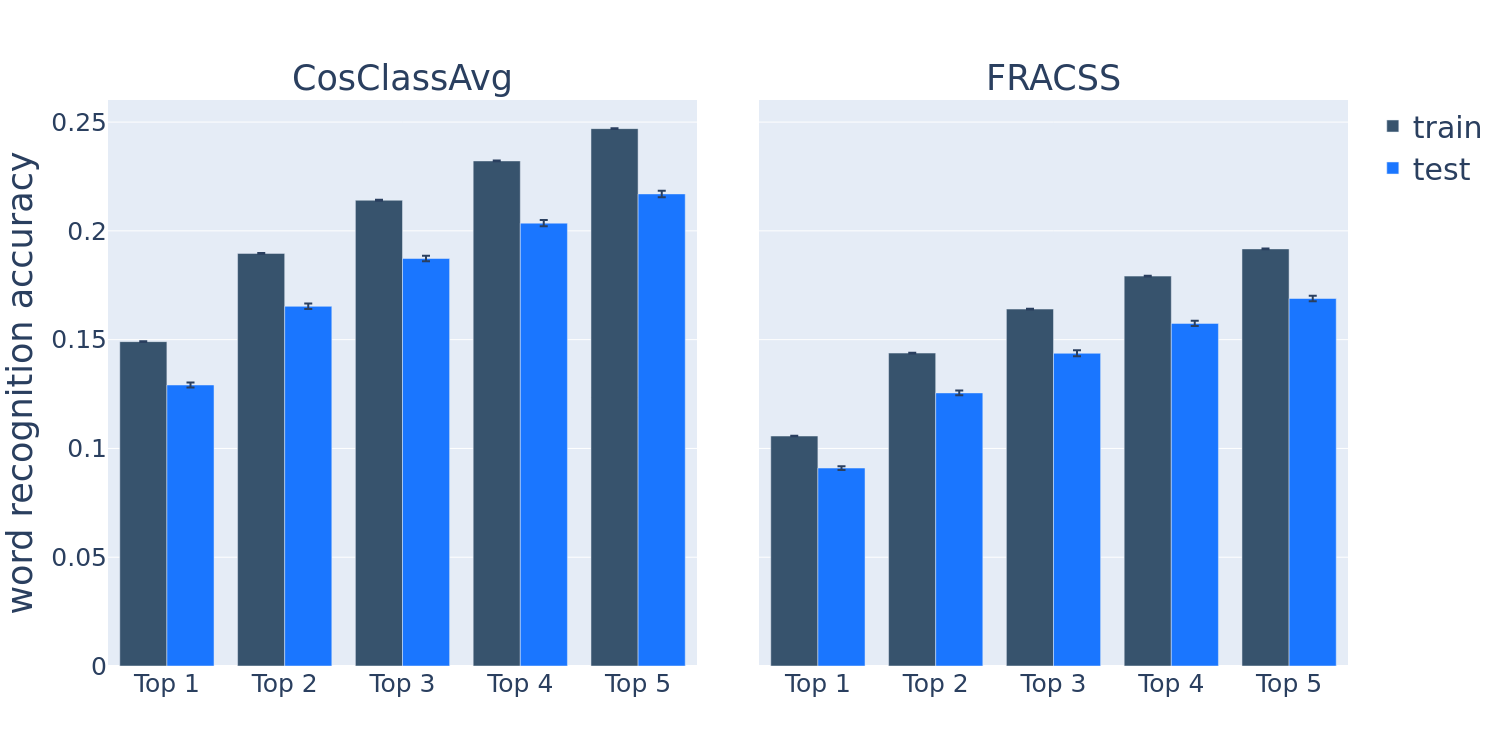}
    \caption{Top-$N$ accuracy of word recognition on the training set (dark blue) and on the test set (light blue) for CCA (left) and FRACSS (right).  The small $1\mathrm{SE}$ error bars indicate that there is very little variability across the 10 cross-validation folds.}
    \label{fig:top5-acc}
\end{figure}

Figure~\ref{fig:top5-acc} presents top-$N$ accuracy (for $N=1, 2, \ldots, 5$) for training data (dark blue) and test data (light blue), for $\Phi_{\textsc{cca}}$ (left) and $\Phi_{\textsc{fracss}}$ (right).  Unsurprisingly, accuracy increases as the criterion for accuracy is relaxed by increasing $N$.  For $N=5$, $\Phi_{\textsc{cca}}$ is successful for up to 25\% of the training tokens and 22\% of the test tokens.  For $N=1$, accuracy on training is 15\% and around 13\% for held-out data. The drop in accuracy for test data as compared to training data is as expected.  What is important is that the reduction in accuracy is relatively small, an indication that the mappings are not severely overfitting the data.

Recognition accuracies are low.  This is unsurprising, for various reasons. First, it is worth noting that standard deep-learning speech recognition systems do not perform well either on the task of isolated word recognition.\footnote{Comparing the performance of LDL-AURIS with the performance of two Automatic Speech Recognition (ASR) systems, DeepSpeech2 and Kaldi, on 98,883 audio tokens from the NewsScape 2016 corpus, \citet{Shafaei:Tari:Uhrig:Baayen:2021} found that pre-trained Kaldi and pre-trained DeepSpeech2 performed at 22\% and 20\%, respectively, while LDL-AURIS trained on far fewer data performed at 12\%.} It is for the recognition of connected speech that deep learning models show truly impressive  performance.

Second, the audio tokens used in the present study have all been cut automatically based on the word boundaries identified during the forced alignment process (see section \ref{sec:data}). There are many instances in which these cut-points do not coincide with the exact word boundaries, even if the identification of such boundaries was possible at all. 

Third, the models are trained and tested on auditory data spoken by numerous speakers, with no prior speaker normalization. The speakers come from different age groups and genders in diverse environments, some of which have  background noise. 

For the purposes of the present study, the relative performance for plural semantic vectors of the $\Phi_{\text{\textsc{fracss}}}$ and $\Phi_{\text{\textsc{cca}}}$ is at the center of interest. A comparison of the left and right panels of Figure~\ref{fig:top5-acc} clarifies that $\Phi_{\textsc{cca}}$ outperforms $\Phi_{\textsc{fracss}}$ for both training and test data.

\begin{figure}[ht]
    \centering
    \includegraphics[width=.95\textwidth]{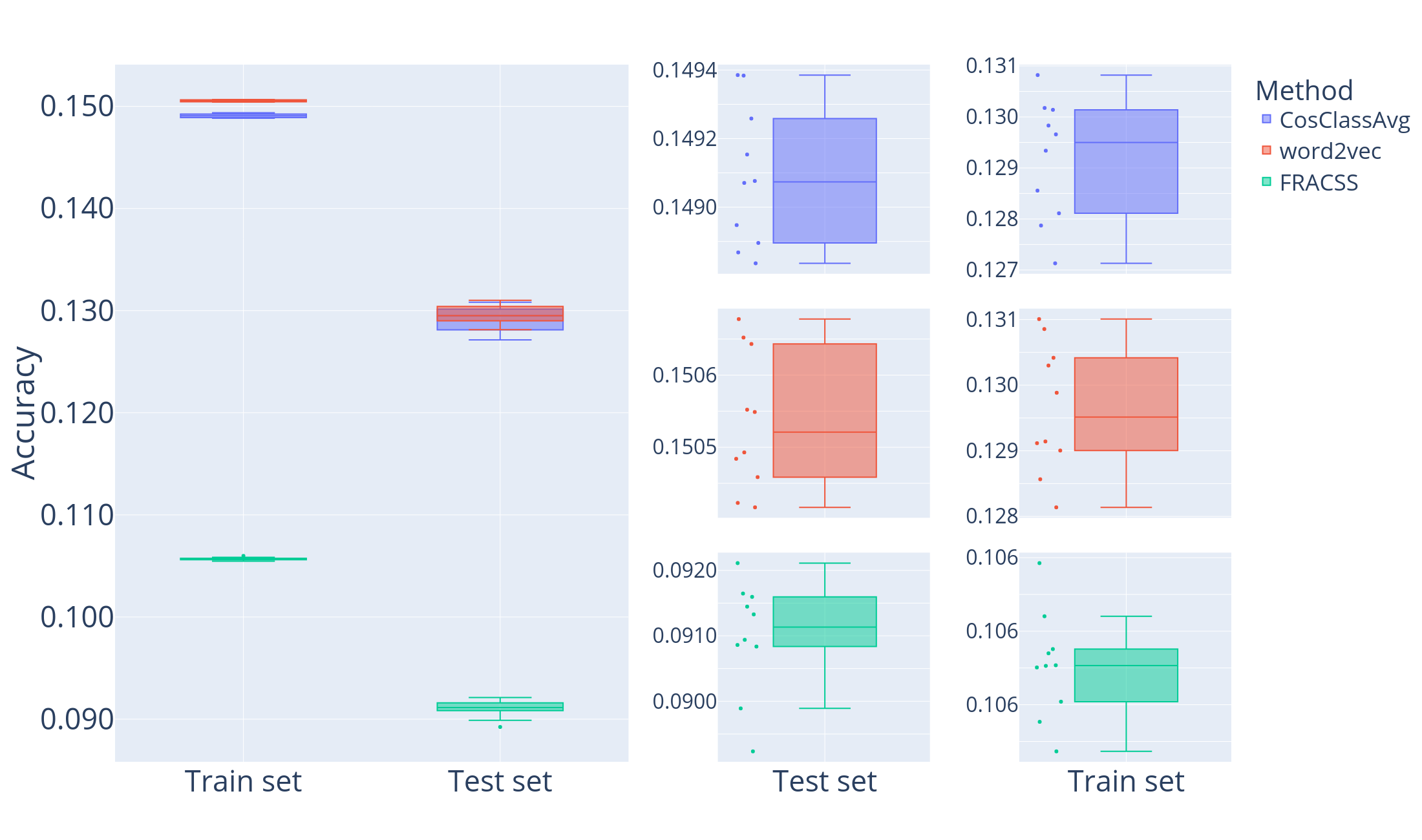}
    \caption{Comparison of the accuracy of  CCA (blue), word2vec  (red), and FRACSS (green), evaluated with a stratified 10-fold cross-validation design on the training and the test sets.  The left panel shows the box plots for all datasets crossed by methods. The panels on the right zoom in on individual plots for improved readability. Each data point represents one fold.}
    \label{fig:acc-all}
\end{figure}

Figure~\ref{fig:acc-all} spotlights the top 1 accuracy measure across 10 cross-validation folds evaluated on the training sets and on the test sets for the three mappings $\bm{F}_{\text{\tiny{CCA}}}$, $\bm{F}_{\text{\tiny{FRACSS}}}$, and $\bm{F}_{\text{\tiny{word2vec}}}$, using box and whiskers plots.  The mapping using CCA reaches a median accuracy of 14.91\% on the training sets and 12.95\% on the test sets, very close to the mapping that uses the semantic vectors provided by word2vec for plurals (15.05\% and 12.95\% for training and test respectively). The mapping using FRACSS plural vectors falls behind the CCA plural vectors by a 4.34 and a 3.83 percentage-point difference in median accuracy on the training and the test sets, respectively. 

\begin{figure}[ht]
    \centering
    \includegraphics[width=.95\textwidth]{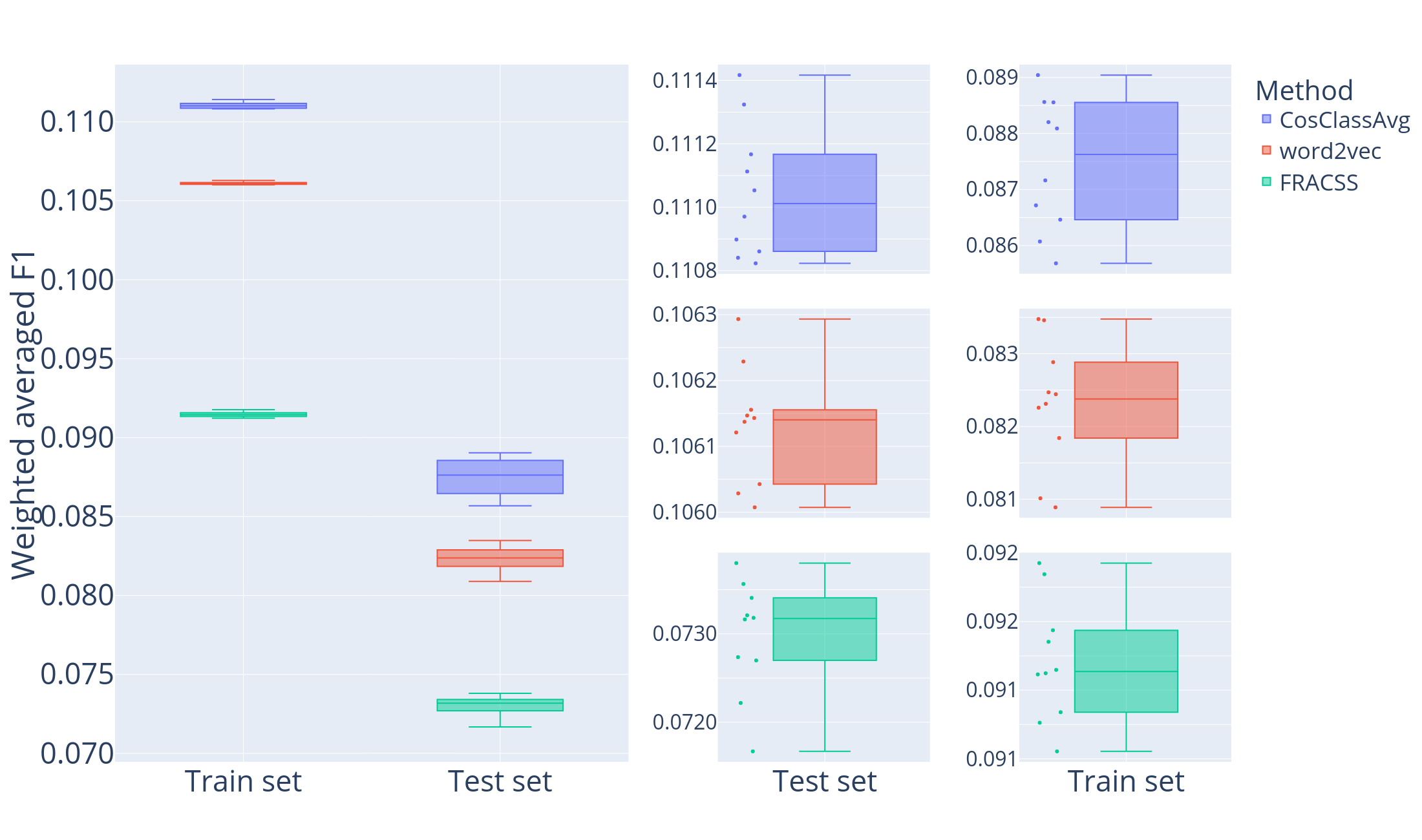}
    \caption{Comparison between the weighted average $F_1$ scores for the mappings using CCA (blue), word2vec (red), and FRACSS (green), evaluated with a stratified 10-fold cross-validation design on the training and the test sets. 
    }
    \label{fig:f1-all}
\end{figure}

As shown in Figure~\ref{fig:f1-all}, when the performance of the mappings is evaluated with weighted average $F_1$ scores,\footnote{A weighted averaged $F_1$ score is the average of the $F_1$ scores for all word types weighted by the number of true instances for each word type.} 
the mapping using CCA outperforms the mapping using word2vec, which in turn outperforms mappings based on FRACSS on the training and the test sets of all folds.


We also compared the accuracy of these mappings with mappings that were set up with semantic matrices for which the row vectors were randomly permuted, breaking the relationship between words' forms and meanings.  As variability in model performance turned out to be very small (see Figure~\ref{fig:top5-acc}), models were trained and tested on just the first fold.  The reduction in accuracy and $F_1$ scores on the training and the test sets are reported in Table~\ref{tab:control_condition}.  Consistent with the previous results, the reduction in performance under random permutation is larger for CCA as compared to FRACSS for all dataset-metric pairs. 

\begin{table}[ht]
    \centering
    \caption{Reduction in accuracy and in weighted averaged $F_1$ score of LDL-AURIS on one cross-validation fold when the relation between form and meaning is broken by randomly permuting the rows of the semantic matrix.}
    \label{tab:control_condition}
    \begin{tabular}{lrrrr} \toprule
        Model & \multicolumn{2}{c}{Reduction in accuracy} & \multicolumn{2}{c}{Reduction in $F_1$ score} \\ \cmidrule(lr){2-3} \cmidrule(lr){4-5}
         & training & test & training & test \\ \midrule
        FRACSS & 0.091 & 0.088 & 0.074 & 0.073  \\
        CCA & 0.135 & 0.126 & 0.093 & 0.088 \\ \bottomrule
    \end{tabular}
\end{table}

\begin{figure}[ht]
    \centering
    \includegraphics[width=\textwidth]{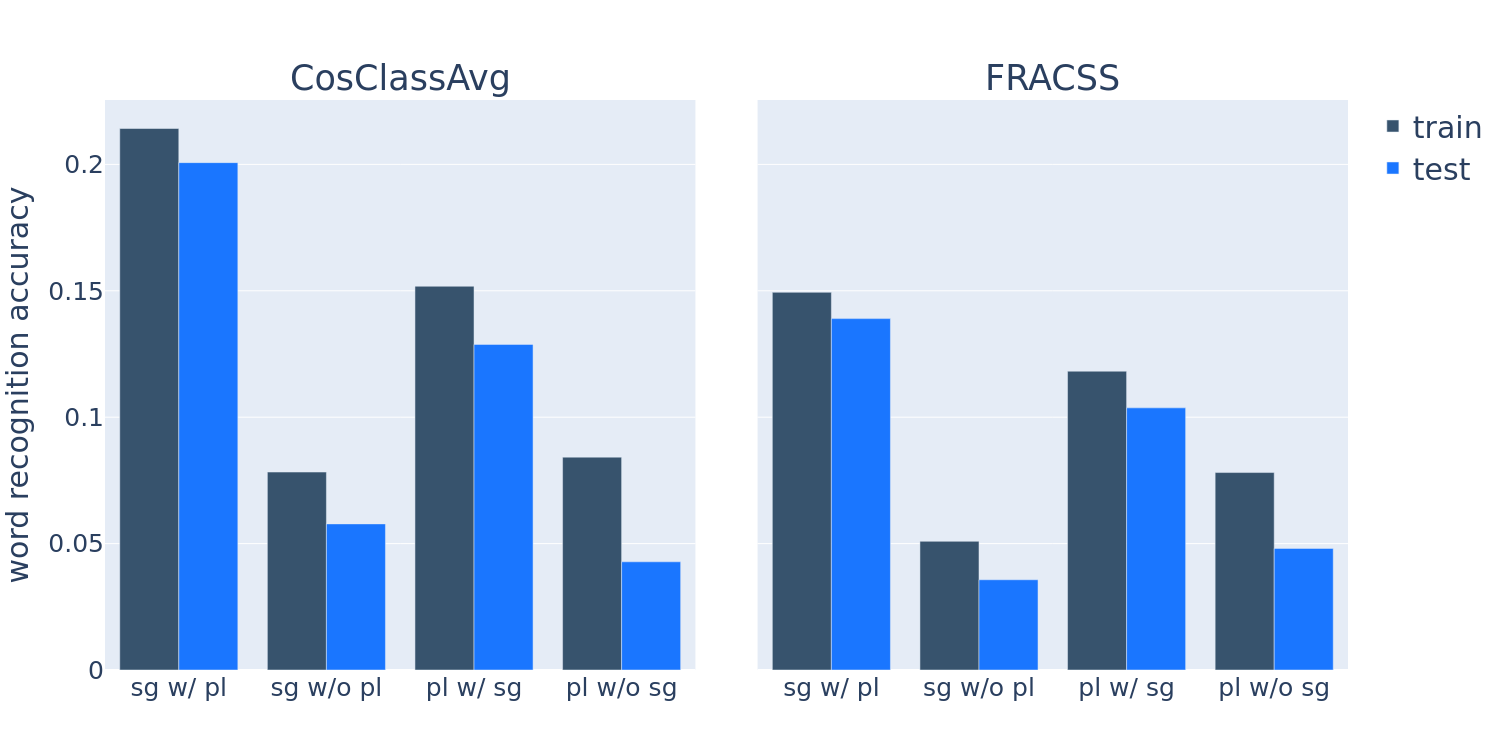}
    \caption{Within-group accuracy for different word groups. In group names, sg stands for \textit{singular}, pl for \textit{plural}, w for \textit{with} and w/o for \textit{without}. A word in the \textit{sg w/o pl} group is a singular word whose plural form is not in the training set. CCA accuracies are greater than those for FRACSS, except for held-out plurals that have no corresponding singulars in the training dataset.}
    \label{fig:acc-category}
\end{figure}

As a final check on the quality of the two implementations of plural conceptualization, we examined accuracy for four subsets of the data, defined by the combinations of the singular and plural forms that were available to the model for training.  Figure~\ref{fig:acc-category} clarifies that accuracy was highest for singular tokens the plural token of which also was encountered during training.  The group with the second-highest accuracy comprises plural tokens that have a singular token in the training set. Accuracies are  substantially lower for singular and plural tokens that were not accompanied by a corresponding plural or singular token during training. (For the counts of auditory tokens within each group, see Table \ref{tab:data} above).  For both conceptualization methods, recognition accuracy benefits from training on tokens of both numbers. Comparing CCA (left panel) with FRACSS (right panel), we can conclude that CCA accuracies are greater than those for FRACSS, except for held-out plurals that have no corresponding singulars in the training dataset.


In summary, higher accuracies and higher higher average $F_1$ scores were obtained with CCA compared to FRACSS, across both training and test data.

\subsubsection{Errors}

\begin{figure}[ht]
    \centering
    \includegraphics[width=0.9\textwidth]{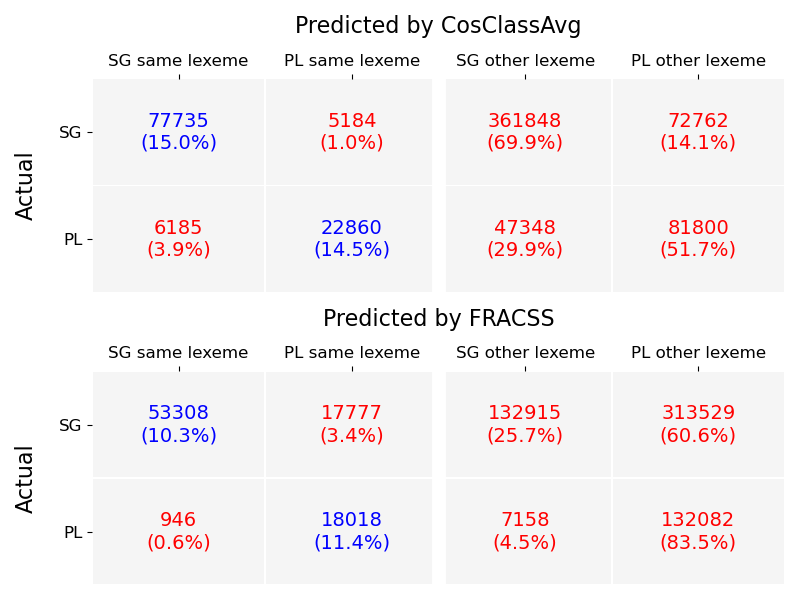}
    \caption{Confusion matrices for the performance of LDL-AURIS with CCA (top) and FRACSS (bottom) vectors on the training set. Correct predictions appear in blue and wrong predictions appear in red.   Percentages are normalized to the sum of rows. $N_{\text{SG}} = 517,529$, $N_{\text{PL}}=158,204$ for both matrices.}
    \label{fig:conf-matrix-train}
\end{figure}

The confusion matrices in Figure~\ref{fig:conf-matrix-train} summarize the predictions of the models for the items in the training set.\footnote{The results on the test set are very similar and reported in the supplementary materials.} Rows correspond to the number feature for the target word: singular words are counted in the first row and plural words in the second row of each matrix. The columns group words by the four combinations of number and whether the lexeme is properly predicted.  Errors are highlighted in red and correct recognitions are highlighted in blue. In addition to the absolute number of tokens, the percentage with regard to the sum of the row is given in each cell. Training with plural vectors obtained with CCA gives rise to larger numbers of correctly predicted target words in the plural as well as in the singular group. 

The two models rarely confuse number when they correctly predict the lexeme.  However, when a lexeme is not recovered correctly, the two models make remarkably different kinds of errors.   For plural vectors generated with CCA, the predicted and targeted word types  tend to match in terms of number.  By contrast, when plural vectors are generated using FRACSS,  the mapping mostly predicts plural forms irrespective of the target word's number. 

Overall, 81\% of the predictions of the mapping using CCA match the target in terms of number.  This percentage reduces to only 50\% for models with plurals generated by FRACSS.\footnote{The estimated difference in proportions between groups at 0.308 is supported by a proportions test (95\% confidence interval of 0.306 and 0.309, $\chi^2(1) = 140897$, $p \ll .0001$).}  Apparently, FRACSS introduces a bias for plural semantics.  This bias may arise due to FRACSS predicting all plurals to be more similar in their semantics, resulting in a strong attractor not only for plurals but also for singulars.  From a methodological perspective, these findings point to the importance of considering predicted plural vectors not just in isolation, but within the full noun system that in addition to plurals also contains singulars.

%

Apparently, plural vectors generated by $\Phi_{\text{\textsc{cca}}}$ are better aligned with information about number in the audio signal than is the case for plural vectors generated by $\Phi_{\text{\textsc{fracss}}}$. This suggests that the global perspective of FRACSS, which takes all singular-plural pairs into account, rather than just those singular-plural pairs that belong to a conceptual class and that share a shift vector, is suboptimal.  

It is surprising that the FRACSS method, which is quite powerful as the underlying linear algebra allows for various geometric modifications such as scaling (stretching or shrinking) and rotation, performs somewhat less well than CCA, a method that from a geometric perspective only implements shifts between vectors.   On the other hand, CCA needs to be informed about semantic classes and assumes that these are given, but setting up these classes is not trivial. 

%

\subsection{An alternative approach}

The models so far were built on the assumption that error vectors $\bm{\epsilon}$ in Equations \ref{eq:cca_e} and \ref{eq:fracss_e} are all measurement noise and reducible. Thus, audio-to-semantic mappings were trained \textit{and} evaluated using predicted plurals by the CCA or the FRACSS conceptualization methods, which build plural vectors without the error term. 

Alternatively, one might assume that error vectors in Equations \ref{eq:cca_e} and \ref{eq:fracss_e} are entirely word-specific semantic information and irreducible. Although we believe, avoiding both of these extremes, the error term comprises measurement noise and word-specific knowledge, the second extreme approach is taken in the following for empirical reasons. Therefore, audio-to-semantic mappings can be trained using the predicted plurals by CCA or FRACSS \textit{but} they should be evaluated using the corpus-extracted vectors, which contain the error term. 

Accordingly, we evaluated the two conceptualization methods by employing the corpus-extracted word2vec embeddings as gold standard, and comparing the gold standard with semantic vectors predicted by the previously obtained form-to-meaning mappings $\bm{F}_{\text{\tiny{CCA}}}$ and $\bm{F}_{\text{\tiny{FRACSS}}}$. For this evaluation, we used the data of the previous cross-validation study. The word2vec vectors are brought together as the rows of a matrix $\bm{S}_{\text{\tiny word2vec}}$. We estimated plural vectors for the test set, and defined a plural vector to be accurate if it was most correlated with the corresponding vector of $\bm{S}_{\text{\tiny word2vec}}$.

\begin{figure}
    \centering
    \includegraphics[width=0.6\textwidth]{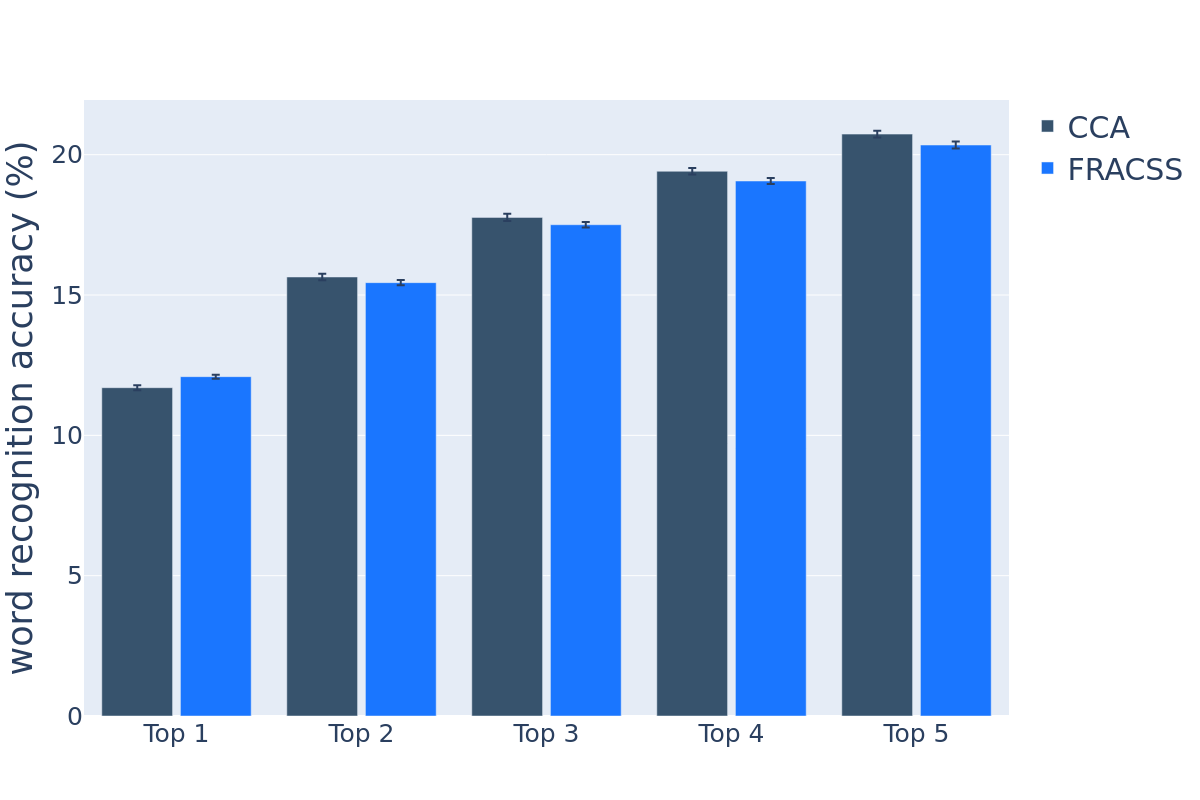}
    \caption{Top-$N$ accuracy (\%) of mappings from audio to semantics, trained either with CCA or FRACSS vectors, but evaluated with word2vec vectors as gold standard. The small $1\mathrm{SE}$ error bars indicate that there is very little variability across the 10 cross-validation folds.}
    \label{fig:extra}
\end{figure}

Accuracies obtained from the second evaluation approach are reported in Figure~\ref{fig:extra}. They are slightly higher for FRACSS for Top 1 evaluation, and slightly higher for CCA for Top 2--5 evaluation. Both of the trained mappings get approximately equally close to the empirical word2vec plural vectors. This suggests that neither CCA nor FRACSS plural vectors are very different from word2vec plural vectors. The results provide further evidence for the observation by \cite{Shafaei:Tari:Uhrig:Baayen:2022:Morphology} that CCA and FRACSS vectors are generally similar to word2vec vectors. In their study, predicted plural vectors by CCA and FRACSS were compared with corpus-extracted plurals using cosine similarity as the measure (a median cosine similarity of 0.71 for CCA and 0.75 for FRACSS). 

These accuracies are within the range of accuracies observed in Figure~\ref{fig:top5-acc}, which were obtained with CCA and FRACSS vectors as gold standard for evaluation. From this analysis, we conclude that the CCA, FRACSS, and empirical vectors are of very similar quality, and that the low word recognition accuracy is not due to low quality of these vectors, but to the enormous variability in the speech signal and the noise in our data.

\section{Study 2: assessing the isomorphy of form and meaning with distance measures}\label{sec:euclid}

We have seen that within the framework of the DLM, semantic vectors conceptualized with $\Phi_{\text{\textsc{cca}}}$ enable a better mapping between words' audio and their meanings compared to $\Phi_{\text{\textsc{fracss}}}$.  In this section, we compare the two conceptualization functions with a method that does not depend on the assumptions of the DLM, such as that the form space and the semantic space can be transformed into each other simply using linear mappings.  

In what follows, we follow in the footsteps of \citet{Monaghan:Shillcock:Christiansen:Kirby:2014},  and investigate possible parallelism between the form space and the meaning space using distance measures. \citet{Monaghan:Shillcock:Christiansen:Kirby:2014} reported that phone-based form similarity (gauged with Euclidean distance and  edit-distance measures) correlated weakly ($r \approx 0.03$) yet significantly with semantic similarity (gauged with the cosine similarity measure between contextual co-occurrence vectors). They observed  correlations for monosyllabic words, uninflected lemmas, and monomorphemic polysyllabic words.

We calculated, for all pairs of words, both phone-based and audio-based measures of their form similarity. In parallel, we calculate the semantic similarity for these pairs of words, using both the original word2vec vectors for plural nouns, as well as plural vectors generated with FRACSS and generated with CCA. For the calculations, the audio tokens from the test set of the first fold of the data (see section \ref{subsec:data}) for which phonemic information was also available were used ($N=60206$, $V=2057$).

First, we calculated the Damerau–Levenshtein edit distance for all pairs of word types in our dataset, which unlike the dataset of Monaghan et al., is restricted to singular and plural nouns. We then calculated the Pearson correlations between these phone-based edit distances on the one hand, and the cosine distances\footnote{The cosine distance zooms in on the angle between two vectors, ignoring differences in length.  The greater the angle between two vectors, the greater their cosine distance is.} between the semantic vectors of these word pairs on the other hand, using three different semantic vectors for plurals: the word2vec plural vectors,  the plural vectors obtained with FRACSS, and the plural vectors obtained with CCA. 


The upper part of  Table~\ref{tab:correlationsFormMeaning} presents the correlation $r$ between the form and meaning distances for all word pairs.  These correlations are smallest for FRACSS, intermediate for word2vec, and greatest for CCA (all $p<.0001$).  The correlations observed for vectors are of the same order of magnitude as the correlations reported by \citet{Monaghan:Shillcock:Christiansen:Kirby:2014}. For a random permutation of the word2vec vectors, the correlation was reduced by a factor of 10 to 0.004.  Our replication study  shows, first, that there is indeed some isomorphism between the form space and the semantic space; and second, that the observations of \citet{Monaghan:Shillcock:Christiansen:Kirby:2014}, which were based on uninflected words, generalize to plural inflection.  Importantly, this analysis shows that, consistent with the results reported in the previous section, CCA-based semantic vectors provide stronger correlations between form distances and semantic distances than FRACSS-based semantic vectors.

\begin{table}[t]
\centering
\caption{Correlations between phone-based and audio-based form distances and semantic distances for three different semantic spaces: CCA, word2vec, and FRACSS. For the audio-based analysis correlations averaged over many samples of audio tokens are reported, with  $\mathrm{SD}$ and the number of trials $N$ in which the correlation was significant at $\alpha = 0.05$.}
\label{tab:correlationsFormMeaning}
\begin{tabular}{lcrrr} \toprule
Form distances & statistic     & CCA & word2vec & FRACSS \\ \midrule
phone-based    & $r$           &  0.061      &  0.041   &  0.038 \\ \midrule
audio-based    & $\bar{r}$     &  0.019      &  0.015   &  0.007 \\ 
               & $\mathrm{SD}$ &  0.003      &  0.003   &  0.005 \\
               & $N$           &   1000      &  1000    &    908 \\ \bottomrule
\end{tabular}
\end{table}

\begin{figure}[ht]
    \centering
    \includegraphics[width=0.8\textwidth]{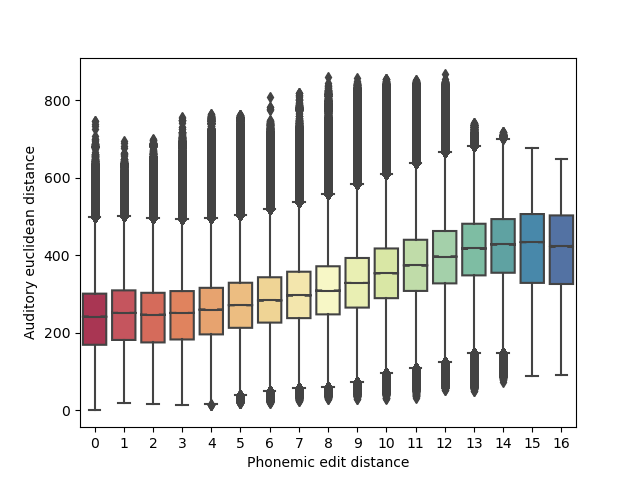}
    \caption{Auditory Euclidean distance as a function of phone-based edit distance.}
    \label{fig:auditory-phonemic}
\end{figure}

For assessing the isomorphy between the form space and the semantic space using the speech signal instead of phone-based representations, we again compared all pairs of words, assessing  their semantic distance with the cosine distance, and evaluated the dissimilarity of their speech signals by means of the Euclidean distance between their C-FBSF form vectors,  which were calculated  from the speech signal as described above in section \ref{sec:repr}.\footnote{By way of example, the list of the top 10 nearest neighbors of an audio token of the word \textit{positions} includes two other audio renditions of the same word form \textit{positions}, one audio token of the singular form \textit{position}, four tokens with similar word-initial phones (\textit{protester}, \textit{process}, \textit{procedures}, and \textit{perspective}), and three tokens with similar sounding word-final syllables (\textit{information} and \textit{conditions}).}  As can be seen in Figure~\ref{fig:auditory-phonemic}, audio-based distance increases with phone-based edit distance ($r(1,812,351,113) = 0.3$, $p < .0001$). As expected, the variability in the audio-based distance as a function of phone-based distance is huge. 

To avoid comparing audio tokens of the same inflectional word type, which is likely to inflate similarity compared to a phone-based analysis in which types do not have non-identical replicates, we randomly sampled for each word type one audio token and then calculated the form distance for each pair of the selected tokens. Subsequently, the correlation between the form distances and the semantic distances of all pairs of inflectional word types was computed. This procedure was repeated 1000 times. The lower part of Table~\ref{tab:correlationsFormMeaning} presents summary statistics for the audio-based comparisons of form and meaning. The average correlations across trials $\bar{r}$, the standard deviations $\mathrm{SD}$, and the number of trials $N$ for which the correlation was significant at $\alpha = 0.05$ are reported. 

Correlations are greatest when semantic vectors are estimated with CCA, and weakest when they are estimated with FRACSS. Randomization of the word2vec semantic vectors resulted in reduced correlations ($M = 0.003$, $\mathrm{SD} = 0.003$). The number of trials in which a significant correlation was observed decreased from 1000 to 744. 

The audio-based correlations are substantially lower than the phone-based correlations. This is unsurprising, as the phone-based analysis do not take into account the huge variability in the speech signal and the highly reduced forms that speakers use in daily speech \citep[see, e.g.,][]{Johnson:2004}.  The small magnitudes of these correlations clarify that evaluations of the similarities between form and meaning run the risk of overestimating these similarities when based on letter-based or phone-based representations. At the same time, it is surprising that significant correlations, however small, can be detected with this method for real, highly diverse, and noisy speech.

Importantly, our audio-based evaluation of the isomorphism between form and meaning provides further support for the possibility that plural vectors conceptualized with $\Phi_{\text{\textsc{cca}}}$ provide a better match with plural forms compared to $\Phi_{\text{\textsc{fracss}}}$. Above, we showed that $\Phi_{\textsc{cca}}$ is optimal for predicting meaning from form, using a linear mapping within the framework of the DLM.  The present observations clarify that this result is unlikely to be an artifact of this linear mapping.  Plural vectors created with $\Phi_{\textsc{cca}}$ better mirror similarities that exist in the form space. 


\section{General Discussion}\label{sec:discussion}

\cite{Shafaei:Tari:Uhrig:Baayen:2022:Morphology} showed that shift vectors, which represent the change in meaning when going from a noun's singular to its plural, cluster by the semantic class of the base word.  Apparently, modeling the semantics of plurality in English with a single shift vector in semantic space, does not do justice to the intricacies of the English noun plural.  Likewise, formal accounts of plurality based on abstract features such as [+\textsc{plural}] lack precision.  From a typological perspective, languages such as Swahili and Kiowa have grammaticalized semantic-class-based differences in their noun classes, languages such as Mandarin Chinese, Japanese and Korean make use of classifiers that are often linked to semantic classes, whereas in English,  systematic differences between words' referents are present only in the semantics as `soft constraints' \citep[see also][]{Corbett:2000:Number}. 

The dependence on semantic class that characterizes the English noun plurals raises the question of how to account for the conceptualization of plurals.  Do we have to construct different plural shift vectors for each semantic class (CCA)?  Or is it possible to predict a noun's plural meaning from a general rule that is informed by all singular and plural nouns in the language (FRACSS)?

\cite{Shafaei:Tari:Uhrig:Baayen:2022:Morphology} addressed this question using the Discriminative Lexicon Model, and observed that semantic vectors for plurals generated with CCA were better aligned with triphone-based form vectors than plural vectors generated using FRACSS.  The present study investigated whether this result generalizes to spoken language. Using the form vectors calculated from the speech signal of a large number of audio tokens extracted from the NewsScape resource,   we observed that the semantic vectors generated by CCA can be predicted with greater precision from their audio than is possible for plural vectors generated by FRACSS.  This holds both for training data and test data.  Furthermore, randomly pairing of form and meaning vectors degrades accuracy more for CCA than for FRACSS.

The superior alignment of the CCA vectors with the form space, compared to FRACSS, received further support from a series of analyses that do not depend on the linear mappings of the Discriminative Lexicon model, and instead follows up on a study by   \citet{Monaghan:Shillcock:Christiansen:Kirby:2014}. For each possible pairing of words in our dataset of singular and plural nouns, we calculated their distance in semantic space as well as their distance in form space (using both phone-based and audio-based distance measures).  Correlations of the distances in form and meaning were stronger for plural vectors obtained with CCA compared to plural vectors obtained with FRACSS.  Considered jointly, we can conclude that approximating the conceptualization of noun plurals in English with CCA likely offers a more precise window on the processing of English noun singulars and plurals in the mental lexicon, the reason being that a better alignment of form and meaning implies greater regularity and hence  facilitation of lexical processing.    

It is surprising that FRACSS does not perform as well as CCA, as the linear mapping that FRACSS makes use of is more powerful than the simple vector addition that CCA employs.  In fact, given the common assumption that pluralization is a unitary operation, a linear mapping as proposed by FRACSS makes sense:  A global singular-meaning-to-plural-meaning mapping is informed by all the singular and plural nouns experienced by the language user, and can be expected to make optimal use of the shared semantic similarities that are supposed to exist between all singular and plural nouns.  However, careful inspection of the distributional semantics of English noun plurals shows that in this language, the semantics of pluralization is not as uniform as previously supposed.  Conceptualization of English plurals is much more locally determined. As a consequence, taking into account how nouns in distant semantic classes are pluralized appears to be detrimental to prediction accuracy.  

When comparing FRACSS with CCA, however, it should be noted that CCA takes for granted that information is available about which semantic class a lexeme belongs to.  As semantic classes are not straightforward to induce in a bottom-up manner --- we made use of WordNet --- CCA is given much more refined information to work with compared to FRACSS.    In terms of model complexity, CCA is therefore likely to be the more complex model. But this increased complexity appears to be justified by greater prediction accuracy.  

We conclude with placing the present study in the wider context of research on spoken morphology.  The data considered in our study are restricted to lexemes that take \{S\} as the exponent for plurality.  Recent studies on English word-final /s/ have clarified that the acoustic duration of /s/ varies systematically with the semantics realized with the /s/ \citep{Plag:Homann:Kunter:2017,Tomaschek:Plag:Ernestus:Baayen:2019}.  In other words, an exponent that in standard theories is assumed to be realized in exactly the same way, independent of its function, is now known to be articulated in ways that reflect its function.  One of the functions of /s/ considered in these investigations is the realization of plurality on nouns.  The present study provides evidence that pluralization for English nouns is not a monolithic and general semantic operation, but varies with the semantic class of the base word. Importantly, we have shown that these differentiated plural semantics are aligned with words' audio signal.  The way in which plurals are articulated therefore depends on the semantics of their base words.  All that we have accomplished in the present study is demonstrate this fact.  In what way these different class-specific semantics are realized in the speech signal is at present unknown, and is a topic for further research.  A better understanding of the phonetic realization of English plurals may further our understanding of why deep learning models in natural language processing and artificial intelligence, which are superbly tuned to the distributional statistics of language use, are so  successful.  This, in turn, may contribute to enhancing models of lexical processing in the mental lexicon.

\subsection*{Acknowledgments}
This research was made possible by funding from the ERC, project WIDE-742545, awarded to RHB, and by funding from the Competence Network for Scientific High Performance Computing in Bavaria to PU. The authors are indebted to two anonymous reviewers, Jessica Nieder, and Melanie Bell for their critical and constructive feedback on our manuscript.

\subsection*{Authors' contributions}
ESB and RHB conceptualized and designed the study. 
PU collected and processed the NewsScape data. 
ESB created the materials in consultation with PU.
ESB implemented the simulations.
All authors were involved in the data analyses and interpretation.
All authors were involved in drafting the manuscript. 

\noindent

\bibliography{bibliography}
\end{document}